%% file: main.tex
\documentclass{article}

\PassOptionsToPackage{numbers, compress}{natbib}

 \usepackage[eandd, final]{neurips_2026}

\usepackage{graphicx}
\usepackage{subcaption}
\usepackage{rotating}
\usepackage{amsmath}
\usepackage{pstricks}
\usepackage{pst-node}

\usepackage{pgfplots}
\usetikzlibrary{arrows.meta}
\pgfplotsset{compat=1.18}

\definecolor{newlocationblue}{HTML}{627EBA}

\usepackage[utf8]{inputenc} 
\usepackage[T1]{fontenc}    
\usepackage{hyperref}       
\usepackage{url}            
\usepackage{booktabs}       
\usepackage{amsfonts}       
\usepackage{nicefrac}       
\usepackage{microtype}      
\usepackage{xcolor}         
\usepackage{cleveref}
\usepackage{enumitem}

\usepackage{xspace}

\makeatletter
\DeclareRobustCommand\onedot{\futurelet\@let@token\@onedot}
\def\@onedot{\ifx\@let@token.\else.\null\fi\xspace}
\makeatother

\title{CFC26: Building Evaluations for Deployment in Sonar-Based Fish Counting}

\author{%
  \bfseries
  Madison Van Horn$^{1}$,
  Suzanne Stathatos$^{1}$,
  Sevan Brodjian$^{1}$,
  Justin Kay$^{2}$,
  Kai Van Brunt$^{2}$, \\
  \bfseries
  Erik Young$^{3}$,
  Sara Beery$^{2}$,
  Pietro Perona$^{1}$,
  Michael Hobley$^{1}$ \\[3pt]
  \mdseries
  $^{1}$California Institute of Technology,
  $^{2}$Massachusetts Institute of Technology, \\
  $^{3}$California Institute of Technology Volunteer \\[3pt]
  \texttt{\{mvanhorn, sstathat, sbrodjia, perona, mahobley\}@caltech.edu} \\
  \texttt{\{kayj, kav, beery\}@mit.edu},
  \texttt{skagitsalmon@gmail.com}
}

\begin{document}

\maketitle

\begin{abstract}
    Accurate counts of fish populations are essential for monitoring ecosystems and making decisions for conservation, harvest limits, and river management.
    In many field deployments, these counts are obtained from sonar videos that experts must inspect manually, making large-scale monitoring slow, expensive, and difficult.
    Automated counting systems have been proposed as a scalable alternative, but so far this promise has been difficult to realize in deployment.
    Developing evaluations that better reflect real-world deployment conditions will help close this gap and drive improvements in counting performance in practice.
    To achieve this, we propose Caltech Fish Counting 2026 (CFC26), a new dataset with improved evaluation protocols that simulate deployment to seen and novel rivers, testing in-distribution and out-of-distribution performance.
    We expand the CFC22 dataset to eleven deployment locations across nine Pacific salmon river systems from Alaska to Northern California, capturing a broader geographical and ecological range of data.
    We additionally introduce two new evaluation metrics for counting, nMNE and nMANE, that mirror the operational workflow that experts use and help surface errors that frame-level detection metrics do not capture.
    Using these metrics, we reveal previously-concealed directional bias in upstream vs downstream counts, which would have direct and significant ecological implications if undetected. The dataset and code are publicly available on \href{https://data.caltech.edu/records/c4y7r-zrc03}{CaltechDATA} and \href{https://github.com/madivanhorn/CFC26}{GitHub}.
\end{abstract}

\section{Introduction}

\begin{figure}[t]
  \centering
  \includegraphics[width=\textwidth]{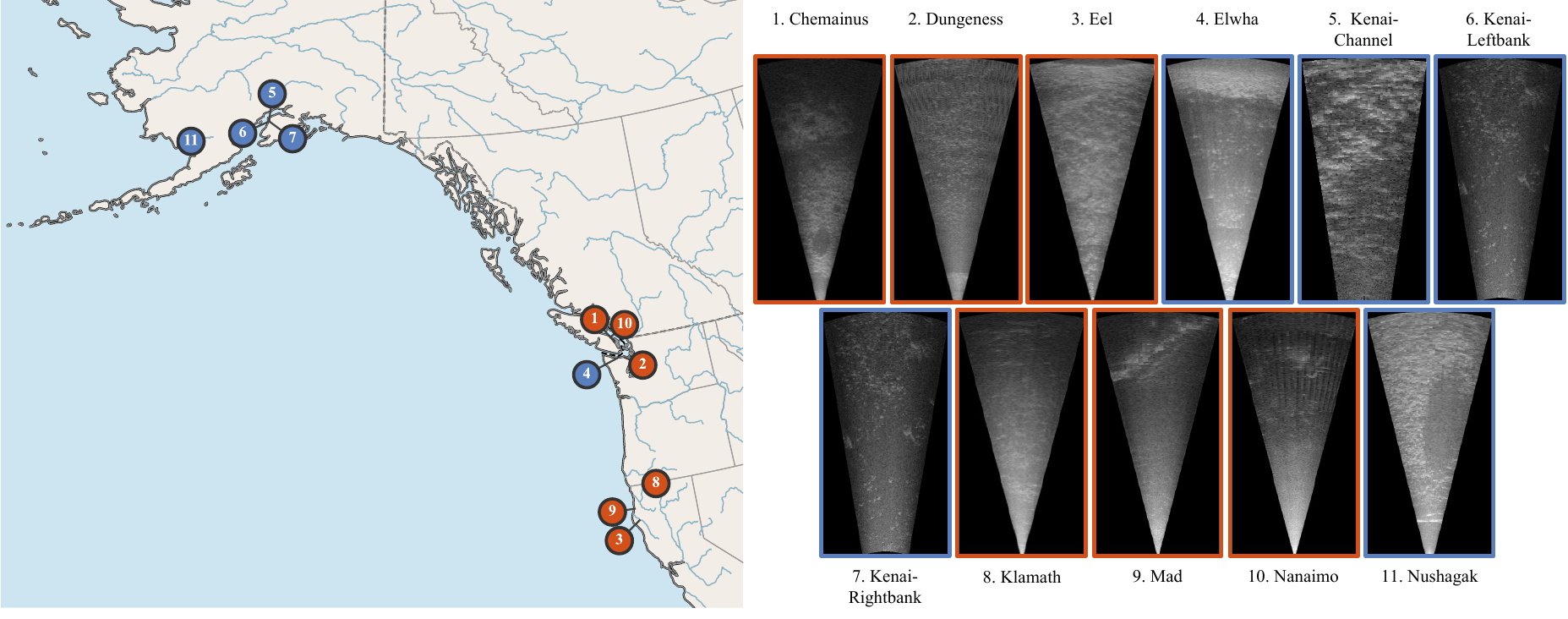}
  \caption{\textbf{CFC26 data is diverse geographically and visually.} Sonar sensors are deployed across nine distinct rivers spanning Alaska, British Columbia, Washington, and California, across eleven deployment locations. Blue markers indicate locations included in CFC22, where we have added additional data and annotations; orange markers indicate new CFC26 locations: (1) Chemainus, (2) Dungeness, (3) Eel, (4) Elwha, (5) Kenai-Channel, (6) Kenai-Leftbank, (7) Kenai-Rightbank, (8) Klamath, (9) Mad, (10) Nanaimo, (11) Nushagak.}
  \label{fig:deployment_map}
\end{figure}

Machine learning models improve via a familiar loop: researchers release a dataset, the field trains models on it, leaderboard progress identifies stronger systems, and new datasets or evaluation protocols reset the target.
This loop has produced real advances by making progress measurable, comparable, and cumulative.
Benchmarks define the task, datasets determine the variation models are exposed to, and metrics decide which failures matter.
As models improve, those same benchmarks often become the evidence that a system is ready to move from research to practice.

But this lifecycle often treats deployment as an endpoint rather than as feedback about what the benchmark failed to measure and how it should be updated.
When practitioners deploy machine learning systems, new failure modes surface: data distributions shift, information arrives sequentially rather than as randomly sampled examples, and operational objectives differ from the proxy metrics used during development.
These failures rarely feed back into how the field measures progress.
As a result, a system can lead a leaderboard while failing in the field.

We have witnessed this gap between benchmarking and practice directly in our own work developing systems to count salmon in sonar video, and see this as a rich testbed to explore the complexities and practical realities of deployment in a real, high-impact application.
Fisheries across the Pacific Northwest deploy sonar sensors to estimate population size for regulation and environmental policy.
The task is hard: sonar video is low signal-to-noise (SNR), individual fish are often indistinguishable from background in any single frame, and the useful signal is motion over time rather than stable appearance.
Models must run over long deployments under widely varying conditions: fish size, density, sensor range, and background statistics all shift with river ecology and geography. Existing workflows with bespoke detection models still are predominantly manual. A model may speed up practitioners identifying true positives, but when models deviate too much from what practitioners expect, all false positives, and tediously, false negatives, are still left to be checked. 

To see methods deployed broadly and address the needs of fisheries managers, we need models that can benefit from training across rivers, generalize reliably beyond the environments on which they were developed, and robustly capture ecological trends without inducing significant bias. 

The Caltech Fish Counting Dataset (CFC22) \cite{kay2022caltechfishcountingdataset} established the primary evaluation protocol for this task (generalizing fish counting to previously unseen rivers), and, while it demonstrated that generalization is difficult and sensitive to ecological variation, it also provided evidence that automated sonar counting was achievable within bounds of error that ecologists would accept.

But when we tried to actually deploy models that performed well on CFC22, these results didn't hold.  
Deployed locations that diverged further from the training river exhibited larger performance drops, indicating a real deployment failure mode that intensifies under broader ecological shift.
The `out-of-distribution' rivers in CFC22 were not ecologically diverse enough to fully stress-test deployment-specific overfitting.

Through deployment, we also discovered a gap between the metrics used to measure counting performance in CFC22 and the operational goals of fisheries managers even for ``high-performing, in-distribution'' rivers.
The counting metric we proposed for CFC22, nMAE, measures absolute crossing error, but is not informative about directional bias or whether errors are from over- or under-counting.
In deployment, the compounding challenges of greater environmental variability  and directional counting bias, were both clear and consequential.
In this work, \textbf{we re-examine our previously developed goalpost for this task through the lens of experience in practical deployment}, identify where its evaluation protocol didn't align with deployment, and redesign it to redefine the goalpost closer to reality, better measuring the failure modes that matter operationally to drive methodological progress.

We introduce CFC26, an expanded benchmark for deployable sonar-based fish counting.
CFC26 expands CFC22 from 1,567 to 4,363 annotated videos and from 515,933 to over 1M bounding boxes across eleven deployment locations spanning nine ecologically distinct river systems in three states and two countries (Figure~\ref{fig:deployment_map}).
In deployment settings, training and evaluation data are not exchangeable random samples: videos are collected over time as river conditions, fish density, and background statistics change. CFC26 mirrors this with temporal data splits.
Combined with the increased difficulty and diversity of data, this makes the benchmark a more robust proxy for deployment to new locations. 

Our contributions are:
\begin{enumerate}[
    label=(\arabic*),
    leftmargin=*,
    align=left,
    itemsep=0.25em,
    topsep=0pt,
    parsep=0pt,
    partopsep=0pt
]
\item \textbf{Revising the evaluation protocol to better match deployment} at varied locations through in-distribution (ID) and held-out-river out-of-distribution (OOD) protocols that make generalization a measurable property and directly simulate the deployment conditions practitioners face;
\item \textbf{Two novel counting metrics}, nMNE and nMANE, that enable the detection of directional bias in aggregate passage counts; and
\item \textbf{A dataset of 4,363 annotated videos and over 1M bounding boxes} across eleven deployment sites spanning nine ecologically distinct river systems, providing the geographic and ecological diversity required to support meaningful OOD evaluation.
\end{enumerate}

\section{Related Work}
\textbf{Large datasets and benchmarks.}
Recent benchmark work in computer vision shows that progress depends not only on scale, but on the interaction of dataset size, composition, and evaluation protocol \citep{gadre2023datacomp, koh2021wilds}. Large-scale datasets such as iNaturalist and Segment Anything illustrate that benchmark expansion is most useful when additional data broadens real-world coverage and the annotation pipeline is aligned with the intended use case \citep{vanhorn2018inaturalist, kirillov2023segmentanything, ravi2024sam2, carion2025sam3segmentconcepts}.

\textbf{Variability measurements in benchmarks.}
Previous works also question and measure variations in benchmarks.
\citet{torralba2011datasetbias} and \citet{recht2019imagenetv2} show that dataset bias and refreshed test sets change benchmark conclusions when evaluation data exposes structure the original benchmark missed. \citet{taori2020naturalshifts} similarly show that robustness gains depend on which natural distribution shifts are evaluated; WILDS made this principle explicit by organizing evaluation around deployment-relevant variation \citep{koh2021wilds}. CFC26 follows this view: its value comes not only from more annotations, but also from exposing data variation encountered at deployment.

\textbf{Imaging sonar datasets.}
ARIS (Adaptive Resolution Imaging Sonar) and DIDSON (Dual-frequency IDentification SONar) are the primary sensing modalities for underwater fish monitoring in turbid river environments. Prior ARIS and DIDSON datasets established the feasibility of sonar-based fish analysis, but most were narrow in scale, task, geography, or annotation type. \citet{mccann2018rawdidsondataset}, for example, released a DIDSON dataset for fish classification with 524 short species-labeled clips from a single river and few taxa. Other ARIS works addressed narrower settings such as single-site dense-school counting \cite{tarling2022deeplearningcountfish,cantor2021underwatersurveysmulletaris} and tailbeat-based species differentiation \cite{helminen2021tailbeatfrequenciesaris}. The closest prior line of work focused on automated salmonid counting in the Kenai River in Alaska: \citet{kulits2020automatedsalmonidcounting} developed an ARIS detector on 527 clips from five Kenai sonar sensors, but split only across sensor placements within one river. CFC22 \cite{kay2022caltechfishcountingdataset} expanded this into a benchmark for detection, tracking, and counting with limited out-of-distribution evaluation. Collectively, these datasets and evaluation benchmarks defined the tasks but did not incorporate sufficient evaluation feedback from the operators to be used in practice -- CFC26 does.

\textbf{Counting}.
Beyond sonar, the broader counting literature shows that benchmark design determines which failures are visible. Benchmarks in crowds, traffic, microscopy, drones, few-shot counting, and underwater imagery have expanded counting across modalities and supervision regimes \cite{UCF-CC-50,worldexpo10,sindagi2020jhu-crowd++,gao2020nwpu,TRANCOSdataset_IbPRIA2015,Marsden_2018_CVPR,dronecrowd_cvpr2021,hobley2022-LTCA,Ranjan-etal-CVPR21,hobley2023abc,sun2023ioc}. However, most still evaluate per-image or per-frame cardinality with MAE, MSE, or localization-style metrics.

For sonar fish counting, that framing is insufficient. Most imaging-sonar salmon-counting methods focus either on frame-wise detection \cite{qian2026self} or on counting via tracks built from frame-wise detections \cite{kay2022caltechfishcountingdataset,kulits2020automatedsalmonidcounting,stathatos2026saved}. At the same time, a growing line of work uses a track-first framing to improve detection \cite{trace}, or temporal representations to directly regress counts \cite{vanbrunt2024Counting} or segment tracks \cite{Tardis}. The operational target, however, is aggregate passage at a previously unseen river, so the relevant task sits at the intersection of detection, tracking, and counting. In this setting, frame-level accuracy matters only insofar as it produces reliable deployment-level counts and exposes aggregation failures such as directional bias.

\textbf{Low-SNR video datasets}
Several adjacent benchmarks emphasize the same core difficulty: appearance is weak and motion carries the signal. Examples include camouflaged-animal datasets \cite{bideau2016itsmoving,lamdouar2020betrayed} and medical video datasets \cite{CVANet,born2021l2,born2021accelerating}. CFC26 belongs to this family, but adds the deployment requirement that temporal signal must support accurate aggregate passage estimates at unseen rivers.

\section{The CFC26 Dataset and Benchmark}
Humans determine net fish passage by counting crossings across a center line in each direction. 

Automated counting systems typically mirror this workflow by detecting fish from frames in video, creating tracks, and enumerating center line crossings.
For these methods to be useful in deployment, the evaluation must indicate how well a system will perform at a new river and whether its errors are aligned with the quantities used for ecological decision-making.
However, current methods often fall short in accuracy, especially at novel locations, and require significant human-in-the-loop correction.

Prior datasets are limited in geographic diversity, temporal coverage, and annotation quality. 
At the same time, the dominant prior metric, nMAE, is not aligned with ecological goals and was not expressive enough to clarify if a method was under- or over-counting. 
These limitations obscure true model performance, particularly under distribution shift. 
CFC26 is designed to address these gaps by emphasizing deployment difficulty and diversity, improving annotations, and introducing count-based evaluation metrics aligned with ecological monitoring objectives.

\paragraph{Diversity and Scale} CFC26 spans eleven deployment locations across nine river systems in three states and two countries (Figure~\ref{fig:deployment_map}). The dataset captures substantial variation of field conditions: fish vary substantially in size and density, and sonar ranges span from short distances (<6m) to long-range deployments ($\sim$50m).  Together, this diversity enables CFC26 to evaluate generalization across ecologically and operationally distinct deployment settings.

In total, CFC26 contains 4,363 video clips, and over 1M annotated bounding boxes (694,273 train, 181,453 val, and 132,151 test) across 14,072 tracks, nearly doubling the scale of CFC22 (1,567 clips, 515,933 boxes). For Elwha and Kenai-Channel, we extend CFC22 with newly annotated clips, 25\% and 63\% extra clips, respectively. Table~\ref{tab:summary_stats} summarizes dataset statistics across all locations.
All footage was captured using stationary ARIS Explorer systems operating at either 1.2 or 1.8 MHz with either 48 or 96 beam configurations. In some cases, a telephoto lens was used to capture long-range imaging.

\begin{figure}[t]
  \centering
  \includegraphics[width=\textwidth]{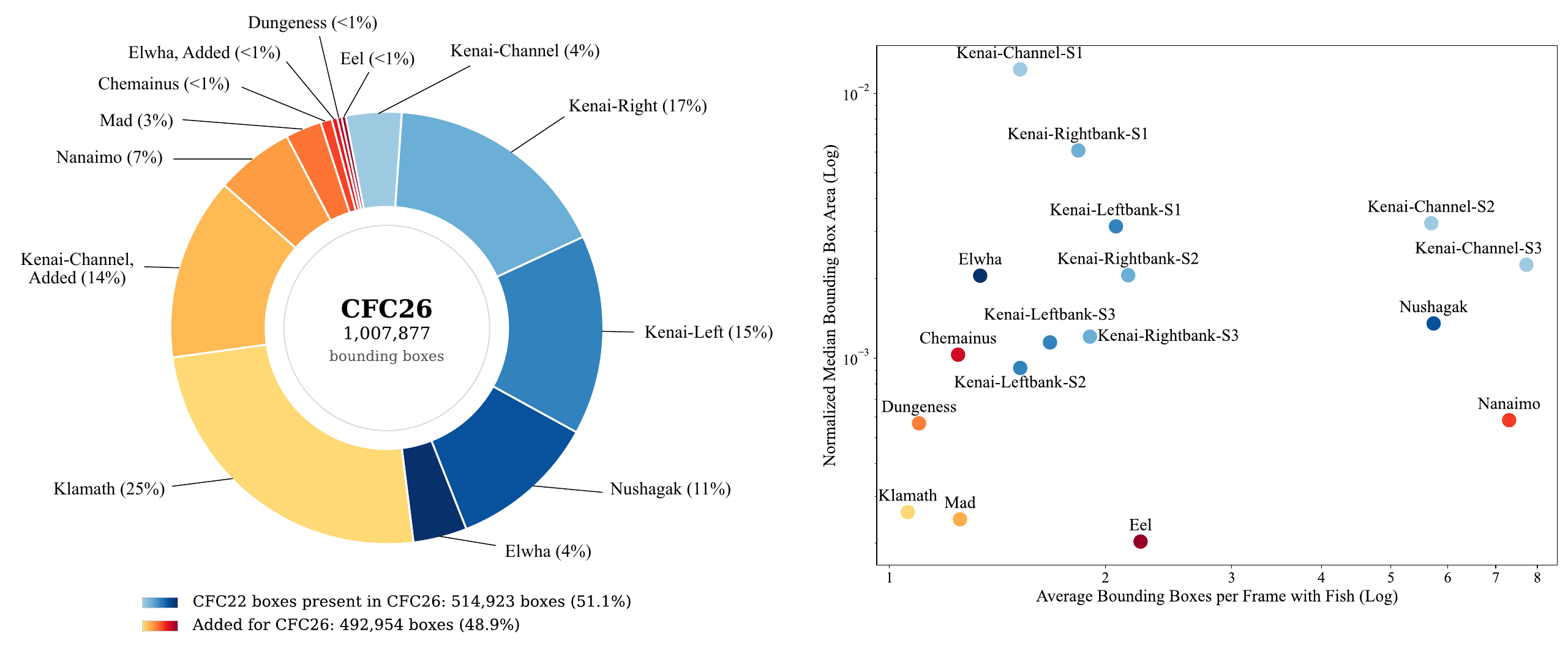}
  \caption{\textbf{CFC26 expands coverage with existing and newly added locations.} The left chart shows the number of bounding boxes per location, highlighting each river’s contribution to the dataset. Elwha and Kenai-Channel include additional annotations not present in CFC22. The right plot shows how locations differ: CFC22 locations generally have higher densities and larger bounding boxes than newly added locations. Kenai locations are shown by stratum rather than aggregated, exposing variation across sonar range configurations within the same deployment site.}
  \label{fig:loc_contributions}
\end{figure}

We select locations to jointly cover the operational range of sonar-based fish counting deployments in Pacific salmon monitoring. As shown in Figure~\ref{fig:loc_contributions}, locations span nearly two orders of magnitude in  fish size (2.03e-04 to 1.23e-02 normalized area in frame) and one order of magnitude in fish density (0.42 to 7.69 fish per frame): California rivers (Klamath, Mad, Eel) feature small, sparsely distributed fish, while Alaskan deployments (Kenai, Nushagak) feature larger, denser populations.

Certain locations provide especially challenging test cases. Klamath, for example, combines small fish with longer sonar ranges of approximately 34 meters, making it one of the most difficult settings. Accordingly, it comprises the largest share of the dataset (53.9\% of clips and 24.8\% of bounding boxes). This diversity makes CFC26 a more robust and discriminative benchmark.

CFC26 is constructed from expert-annotated fish crossings within continuous sonar recordings. For each annotated crossing, we identify the frame in which the fish crosses the counting line and extract a temporal window around that frame to form a video clip. Each clip therefore contains at least one annotated fish crossing, although additional fish may also appear and cross within the same temporal window. Temporal padding varies across locations due to historical dataset construction. Frames without fish are included within these clips to provide a realistic temporal context. Negative-only images are excluded during detector training.

\begin{figure}[t]
  \centering
  \includegraphics[width=\textwidth]{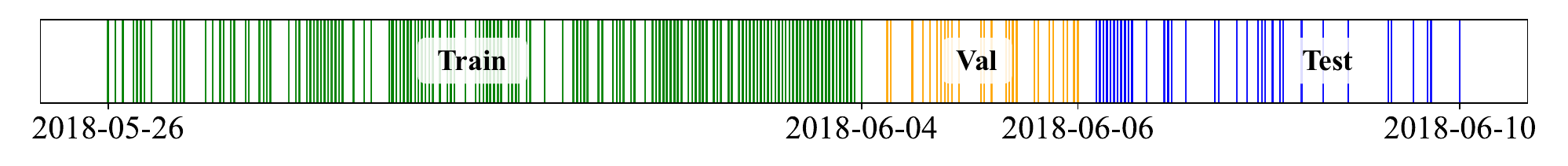}
  \caption{\textbf{Example of temporal distribution of annotated clips for Kenai-Leftbank Stratum1.} Data splits were constructed by identifying gaps in coverage to define temporally separated train, validation and test splits, reducing any potential data leakage. This was applied across all locations.}
  \label{fig:temporal_splits}
\end{figure}

\paragraph{Temporal Coverage and Split Construction}
In real deployment scenarios, models are applied to future, unseen data. CFC26 reflects this with 
splits that are constructed using natural river-specific temporal gaps between clips to form chronologically ordered, non-overlapping training, validation, and test sets (Figure~\ref{fig:temporal_splits}).
We target splits of approximately 60\% training and at least 10\% each for validation and test, with exact proportions determined by deployment duration and gap structure.

\paragraph{Annotation Methodology and Quality Assurance}
High-quality annotations are critical for reliable evaluation. Analysis of CFC22 revealed substantial label omissions, particularly in the Kenai-Channel subset.
Re-annotation of this subset recovered 55.1\% additional tracks, indicating that prior annotations significantly undercounted fish passage.
Such errors can distort model comparisons in subtle but consequential ways \cite{northcutt2021pervasivelabelerrorstest}. 

CFC26 adopts an improved annotation pipeline to ensure completeness and consistency.
Annotators are provided with MP4 videos of the raw video and an echogram visualization \cite{Tardis}, which collapse temporal sonar data into 2D representations where fish appear as distinct trajectories.
This matches the tools used by domain experts \cite{arisfish,echotastic} and is particularly effective for identifying small or low-contrast fish.
Echograms are used only as an aid for detection and do not alter annotation definitions.

All annotations undergo internal quality control using CVAT's annotation platform \cite{cvat}. Videos with annotation errors are re-annotated when possible. Videos are excluded when annotations or tracks deviate substantially from the expected quality, disagreements between reviewers cannot be confidently resolved, or a substantial number of annotations are missing from either the original Caltech Fish Counting Dataset \cite{kay2022caltechfishcountingdataset} or newly completed annotation jobs. In total, 2,794 annotated videos were excluded during this process, primarily due to annotation quality rather than inherent differences in scene difficulty. These quality-control measures substantially improve annotation fidelity, making CFC26 a cleaner and more reliable benchmark.

CFC26 is hosted on CaltechData with publicly available, versioned releases. Future annotation corrections or updates will be incorporated into new dataset versions, preserving prior releases to ensure that experimental results remain reproducible.

\begin{table*}
\centering
\caption{\textbf{CFC26 scales and diversifies CFC22.} Our dataset includes 2.5$\times$ as many clips, 2$\times$ as many frames, and 2$\times$ as many bounding boxes than CFC22, and \textcolor{newlocationblue}{6 additional locations} that expand OOD evaluation to more realistically reflect deployment needs.}
\resizebox{\linewidth}{!}{%
    \input{Tables/dataset_summary_stats_1}
}
\label{tab:summary_stats}
\end{table*}

\subsection{Benchmark Evaluation Protocol and Metrics}
Careful evaluation design is essential if benchmark performance is to be informative about deployment. In CFC26, this requires both evaluating whether a method transfers to a river it has never seen and measuring performance using metrics that reflect the quantities practitioners actually care about. Accordingly, \cref{sec:novel_location_generalisation} defines our deployment-oriented generalization protocol, and \cref{sec:counting_metrics} introduces the counting metrics we use to evaluate method performance under that protocol.

\subsubsection{Generalization Evaluation}\label{sec:novel_location_generalisation}
Our evaluation is designed to answer one deployment-critical question directly: \textit{will this system work at a river it has never seen?} To do so, we evaluate each method under two training regimes.

\textbf{Out-of-distribution (OOD, held-out river):} Entire locations are excluded during training and used only for evaluation. This is the primary regime in CFC26, because it directly simulates first deployment at a new river and therefore serves as our main measure of generalization.

\textbf{In-distribution (ID):} Training and testing include the same location(s), measuring performance when labeled data from the deployment location is available.

OOD performance is informative on its own, but is most useful when interpreted alongside ID performance. Strong ID performance paired with weak OOD performance indicates that a method can fit known locations but does not transfer reliably to new ones. The size of the ID--OOD gap helps identify where ecological or sensing differences between rivers are not captured by the training distribution. Conversely, if both ID and OOD performance are poor, the limitation is less likely to be generalization alone and more likely to reflect insufficient training coverage, limited signal quality, or intrinsic difficulty at that river.

\subsubsection{Counting Metrics}\label{sec:counting_metrics}

To better align evaluation with ecological objectives, we focus on counting-based metrics rather than frame-level detection metrics.
Our primary objective is accurate fish passage estimation, which depends not only on detecting fish in individual frames but also on how detection errors propagate through tracking and crossing estimation.
Frame-level metrics such as mAP50 are therefore not reliable proxies for counting performance: missed detections near track boundaries or low-confidence false positives can fragment or inflate trajectories, producing large counting errors even when detection quality appears high.
Conversely, moderate detection performance can still yield accurate counts when tracking recovers complete trajectories.
For this reason, we use mAP50 only for model selection and report counting-based metrics as the primary evaluation criteria.

Our counting protocol mirrors the operational workflow used by fish biologists by evaluating aggregated crossing counts over a deployment window.
Each video captures fish crossing a virtual center line in the image: fish are first annotated with frame-level bounding boxes, annotations are linked into tracks, and counts are obtained from track crossings of this center line.
We include metrics that separately capture total error and directional bias, since fish passage is fundamentally a signed-flow problem in which symmetric upstream and downstream errors can obscure systematic failures.
The upstream direction varies by deployment, we report counts as upstream and downstream rather than left and right to better align with the scientific questions of interest to ecologists.

For a set of tracks $\mathcal{T}_i$ in clip $i$, we define the upstream count, $z^{i}_{up}$, as:
\begin{equation}\label{eq:z_up}
    z^{i}_{up} = \sum_{j \in \mathcal{T}_i}\text{max}(\sum_{t \in T_i} v_{up}(j,t), 0)
\end{equation}
where $T_i$ is the set of frames in clip $i$ and $v_{up}$ indicates whether track $j$ crosses the center line between adjacent frames:
\begin{equation}\label{eq:v_up}
    v_{up}(j,t) = \begin{cases}
+1 & \text{if track $j$ crosses upstream between frames $t-1$ and $t$}\\
-1 & \text{if track $j$ crosses downstream between frames $t-1$ and $t$}\\
0  & \text{otherwise.}
\end{cases}
\end{equation}
The max removes non-upward movement. The downstream count $z^{i}_{down}$ is defined similarly. Predicted counts $\hat{z}^{i}_{up}$ and $\hat{z}^{i}_{down}$ follow the same definitions applied to tracker output.

Given these counts, the per-clip absolute count error $E_i$ and net count error $D_i$ are:
\begin{equation}\label{eq:error_terms}
\begin{aligned}
\mathbf{E_i} &= \left|z^i_{up} - \hat{z}^i_{up}\right| + \left|z^i_{down} - \hat{z}^i_{down}\right|,
&
\mathbf{D_i} &\stackrel{\pnode{E1}}{=} (\hat{z}^{i}_{up} - \hat{z}^{i}_{down}) - (z^{i}_{up} - z^{i}_{down}) \\[0.6em]
&&
    &\stackrel{\pnode{E2}}{=} (\hat{z}^{i}_{up} - z^{i}_{up}) - (\hat{z}^{i}_{down} - z^{i}_{down}).
\end{aligned}
\end{equation}
$E_i$ sums the unsigned error in each crossing direction; $D_i$ measures net directional bias, with the second line decomposing it into per-direction errors.

We then aggregate these per-clip quantities into three counting metrics:
\begin{equation}\label{eq:error_metrics}
\begin{gathered}
\text{\textbf{nMAE}} = \frac{\sum_{i=1}^{N}E_i}{\sum_{i=1}^{N} \left(z^{i}_{up} + z^{i}_{down}\right)}, \qquad
\text{\textbf{nMNE}} = \frac{\sum_{i=1}^{N}D_i}{\sum_{i=1}^{N}\left(z^{i}_{up} - z^{i}_{down}\right)} \\
\\
\text{\textbf{nMANE}} = \frac{\sum_{i=1}^{N} \text{sign}(z^{i}_{up} - z^{i}_{down}) \cdot D_i}{\sum_{i=1}^{N}|z^{i}_{up} - z^{i}_{down}|}, \quad
\text{where } \text{sign}(x) = +1 \text{ if } x \geq 0,\; -1 \text{ otherwise.}
\end{gathered}
\end{equation}

\noindent\textbf{nMAE} aggregates absolute error over all clips, normalized by total ground-truth crossing volume. It was originally proposed in \citet{kay2022caltechfishcountingdataset}; we adopt the same definition, with $z_i$ denoting ground-truth and $\hat{z}_i$ predicted counts. Note that the original paper uses the inverse convention. nMAE captures total error magnitude but treats over- and under-counting symmetrically, which can obscure bias.

\noindent\textbf{nMNE} aggregates the signed net error across clips, normalized by total net ground-truth movement. Positive values indicate systematic overestimation of net movement; negative values indicate underestimation. nMNE aligns directly with what ecologists report, the net passage count, but errors in opposite directions can cancel, so a low magnitude nMNE does not necessarily imply accurate predictions.

\noindent\textbf{nMANE} addresses this by signing $D_i$ relative to the direction of true net movement per clip. A negative nMANE indicates systematic undercounting in the dominant direction of fish movement, while a positive nMANE indicates overcounting. Values near zero indicate accurate directional counting. Because passage estimates depend on quantifying dominant movement correctly, negative nMANE is the most practically consequential failure mode.

\section{Experimental Setup}

The experiments are designed to answer one core question: how much does performance degrade when a model has never seen the target river? The following conditions vary target-river access from full inclusion to complete exclusion from the training set.

\textbf{Balanced Data Sampling}.
Training is fixed to 10k samples per location, drawn from frames containing fish; locations with fewer than 10k qualifying frames are upsampled. Validation uses the same procedure with 3k samples per location. This condition controls for CFC26’s pronounced location imbalance. Klamath alone contributes 24.8\% of bounding boxes, so balanced sampling helps ensure that per-location results reflect model accuracy rather than sampling frequency. See Appendix~\ref{sec:sampling_ablation} for more  details on sampling strategy selection.

\textbf{ID, OOD, Target River (ID)}.
To test out-of-distribution performance degradation, we trained a model on the target river only (Target River, ID), on all rivers (ID), and on all rivers except the target river (OOD). OOD is CFC26's primary evaluation regime to directly simulate first deployment at an unseen location, with no target data available.
Target River (ID) measures how well a model performs when the deployment goal is a single target location, without relying on data from other rivers. Rivers where performance drops significantly indicate that the model struggles when trained and tested in isolation on that location, likely due to limited training scale, insufficient within-location diversity, or intrinsic difficulty at the river.

\textbf{Detector and Tracker}.
We use YOLOv5-medium \cite{ocher2022yolov5} and ByteTrack \cite{zhang2022bytetrackmultiobjecttrackingassociating}. We selected the detector architecture using mean Average Precision at an IoU threshold of 0.5 (mAP50) on the validation set. We use SGD with an initial learning rate of 0.01 to a final learning rate of 0.0001, momentum of 0.937, and weight decay of 0.0005. Additional details regarding model selection, training, and tracker hyperparameters are included in Appendix~\ref{sec:model_ablation}.

\section{Results}

\begin{figure}[]
  \centering
  \includegraphics[width=\columnwidth]{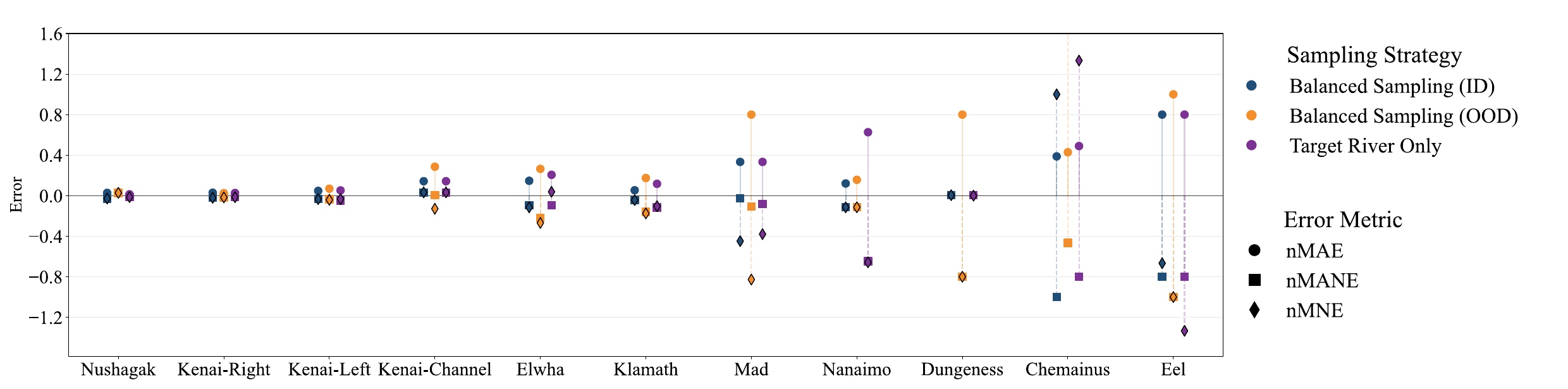}
  \caption{\textbf{Counting errors per location across training conditions.} Balanced Sampling (ID) achieves the lowest nMAE at most locations, while holding out the target     
  river degrades performance substantially at harder sites such as Eel and Dungeness. Directional errors (nMANE) are negative across nearly all conditions, indicating consistent undercounting of the dominant movement direction; Chemainus and Eel show the largest directional bias. Note, the nMNE value for Chemainus under Balanced Sampling (OOD) is a large outlier (3.67) and is truncated from the figure for readability.}
  \label{fig:errors}
\end{figure}

\looseness=-1\paragraph{Target-location data improves performance, but only where the underlying detection problem is tractable.} Balanced Sampling (ID) is the strongest overall condition, achieving the lowest or near-lowest nMAE at most locations (Figures~\ref{fig:errors}). This indicates that broad training diversity combined with target location representation is generally beneficial. The clearest gains occur at locations where errors are moderate rather than severe. For example, Chemainus improves from an nMAE of 0.43 under Balanced Sampling (OOD) and 0.49 under Target River (ID) to 0.39 under Balanced Sampling (ID). Similarly, Elwha remains in a relatively low-error range across conditions, with ID nMAE near 0.15.

In contrast, some locations remain difficult even when included in training. At Eel, Balanced Sampling (ID) still produces nMAE of 0.80, and several other conditions reach the maximum observed error of 1.0. This suggests that performance is not primarily limited by domain shift, but instead reflects inherent difficulty in detecting fish under the given sensing conditions. For Eel, it is not simply an out-of-distribution generalization problem: the small fish size and long sonar range likely make detection fundamentally more challenging there. As a result, adding target location examples improves distributional coverage but does not resolve the underlying visual difficulty.

Mad exhibits a different behavior. Balanced Sampling (OOD) produces a high nMAE of 0.80 while Balanced Sampling (ID) reduces the error to 0.33. Unlike Eel, training on target location data substantially improves performance. This contrast separates two kinds of failure: distribution-sensitive performance degradation versus limitations that persist even with in-distribution training.

\paragraph{Directional error reveals failures that nMAE hides.} Across locations, nMANE is negative in most conditions, indicating a tendency to undercount fish moving in the dominant direction (Figure ~\ref{fig:errors}). This directional bias is especially large at Chemainus and Eel. At Chemainus, nMANE ranges from $-0.47$ to $-1.0$ across conditions, even though nMAE values are moderate. At Eel, nMANE reaches approximately $-0.80$ under Balanced Sampling (ID), $-1.0$ under Balanced Sampling (OOD), and $-0.80$ under Target River (ID). These values show that the model is not merely making noisy count estimates; it is consistently missing directional movement.

The contrast between nMAE and nMANE is clearest at Chemainus. Balanced Sampling (ID) gives the best nMAE at this location, 0.39, but it also produces the worst directional error, with nMANE at $-1.0$. A deployment decision based only on nMAE would therefore select the condition that appears most accurate in absolute count terms, while missing the fact that it severely undercounts the dominant movement direction. This demonstrates why nMAE is insufficient on its own: absolute count error can appear moderate even when directional estimates are strongly biased.

Mad illustrates the opposite case. Under Balanced Sampling (OOD), Mad has high nMAE of 0.80, but its nMANE is much smaller, $-0.11$. This suggests that the model makes substantial absolute counting errors, but that these errors are more balanced across movement directions. For downstream ecological applications concerned with net directional passage, this is a less severe failure than a large one-sided bias, because over- and under-counting in different directions partially offset in the final net estimate.

The Kenai locations are the most stable across both metrics. Kenai-Left and Kenai-Right have low nMAE across all training conditions, generally below 0.07 and 0.03 respectively, and nMANE remains close to zero. These locations thus represent the easiest deployment regime in the analysis.

Overall, the results separate three regimes of behavior. Some locations, such as Kenai-Left, Kenai-Right, Elwha, Klamath, and Nushagak, are robust across training conditions. Others, such as Mad and Dungeness, are strongly sensitive to training distribution, with performance improving substantially when training locations are included. Finally, Eel and Chemainus expose deeper failure modes: Eel remains difficult even with target-location data, while Chemainus appears acceptable under nMAE but fails directionally under nMANE. These regimes closely mirror the conditions practitioners face at new deployment locations, and each requires a different kind of evaluation to surface: held-out-river conditions expose distributional failure, while directional metrics expose counting bias that magnitude error alone cannot detect.

\section{Limitations}  
Our benchmarking on CFC26, while comprehensive, has several limitations. All experiments are reported from single runs without variance estimates or error bars as full multi-seed evaluation across multiple conditions and eleven locations was computationally impractical within this work. The results should therefore be interpreted as directional rather than definitive. Clip counts also vary substantially across locations. Klamath alone contributes 53.9\% of all clips, while several locations have significantly fewer. The Balanced Sampling condition only partially compensates for this imbalance.

CFC26 is also limited to sonar-based fish counting in riverine environments and may not generalize to other sensing modalities (e.g., optical video), marine/open-ocean settings, or species with substantially different motion or morphology. While expanded to nine river systems, geographic and ecological coverage remains incomplete (North America only), and certain river conditions (e.g., extreme turbidity, rare seasonal events, or atypical flow regimes) may be underrepresented. Annotation quality may be affected by low signal-to-noise ratios and ambiguity in fish visibility, particularly in dense or occluded scenes. The dataset is not recommended for tasks requiring fine-grained species classification, individual fish re-identification, or appearance-based recognition independent of motion cues. Finally, CFC26 is a static release, not a continuously updated data engine. Unlike data engines that continuously generate new annotations as models improve, CFC26 is a snapshot of current deployment conditions. This motivates moving beyond a static benchmark-leaderboard cycle toward a deployment-driven lifecycle in which evaluation frameworks are continuously refined to reflect real-world conditions.

\section{Conclusion}
The machine learning development lifecycle is just that, a cycle. We design, test, learn, and redesign, moving ever-closer to our goal of building reliable AI-enabled systems. Evaluations evolve by necessity as part of that cycle, and this work is an example of this evolution. CFC22 drove initial progress, built confidence in the feasibility of the task, and enabled the initial identification of successful methods to improve salmon counting in sonar video. But CFC22, like any benchmark, was not a full snapshot of reality. By deploying methods directly in the field, we learned the hard way what components of our initial evaluation were most unrepresented. This work represents our effort to capture these outstanding gaps in a new snapshot, which moves closer to real deployment in the ways that matter most, and redefine the goalpost for the ML community to ensure that methodological progress towards that goal translates as well as possible to the real needs on the ground.

Towards that end, we expanded our benchmark to cover nine rivers, with nearly 3x the number of video clips giving dense temporal coverage, and a total of more than a million high-quality bounding box annotations, associated to individual tracks, across the data. Scaling our dataset highlights distribution-sensitivities across sonar sensor deployments. We measure gains from additional training data, compared to entirely out-of-distribution setups, in a way that aligns with how practitioners make decisions.
We also introduce new performance metrics that enable us to better capture the ecological implications of models in deployment, such as directional bias in counts leading to over- or under-estimation of fish passage. For example, practitioners optimizing for nMAE would select the configuration that most consistently inverts the dominant crossing direction. nMANE makes this visible; nMAE does not.

Overall, CFC26 represents an important step in the ML deployment \textit{lifecycle}, and one that should not be overlooked. Our evaluations must be as carefully designed, tested, and improved upon as our methods, and true progress is not possible unless we push for improvement in both.

\begin{ack}

We gratefully acknowledge funding from the Caltech Resnick Sustainability Institute Impact Grant “Continuous, accurate and cost-effective counting of migrating salmon for conservation and fishery management in the Pacific Northwest.” We also thank Damiano Marsili and Ilona Demler from Caltech; Keith Denton, a sonar consultant from Washington State, and Alex Stefankiv; and Damon Goodman, James Whelan, and Andrew McLane from California Trout for their valuable feedback and moral support.

\end{ack}

\clearpage
\newpage

\bibliographystyle{plainnat}
\medskip
{
\small
\bibliography{bib}
}

\clearpage
\appendix

\section{Technical appendices and supplementary material}

\subsection{Model Selection and Experimental Setup}
\label{sec:model_ablation}

\paragraph{Detector Architecture Selection.}
We benchmarked four YOLO variants: YOLOv5s, YOLOv5m \cite{ocher2022yolov5}, YOLO26s, and YOLO26m \cite{sapkota2026yolo26keyarchitecturalenhancements}. Each model was trained on the CFC26 training set for two epochs and evaluated on the validation set. We chose this training duration based on early performance saturation: mAP50 plateaued quickly, and additional training yielded negligible gains. The final model was selected based on validation-set mAP50. 

Improvements in detection quality generally translate to improved downstream performance in tracking and counting. Using detection performance as a proxy for end-to-end system performance is substantially more computationally efficient than evaluating the complete detection, tracking, and counting pipeline for every candidate model. However, frame-level detection performance does not necessarily translate to accurate fish counts, as tracking errors, fragmented tracks, and duplicate identities all contribute to counting error. Occasional missed detections may have little effect on the final count if the track remains recoverable. Since fisheries practitioners ultimately care about fish passage estimates, we emphasize the proposed counting metrics because they more directly reflect deployment performance and practitioner priorities while mAP50 serves as an efficient metric for architecture selection. The ranking under mAP50 is generally consistent with the rankings under the counting metrics.

\begin{table}[ht]
\centering
\caption{Model performance comparison}
\label{tab:attr3}
\input{Tables/architecture_selection_3}
\end{table}

\paragraph{Detector Architecture Robustness.}
To evaluate whether the benchmark conclusions depend on the choice of detector architecture, we additionally train RT-DETRv2-S, a transformer-based detector with comparable model capacity and computational complexity to YOLOv5m (20.0M vs. 21.2M parameters; 60 vs. 49 GFLOPs). Despite the architectural difference, the two detectors exhibit similar overall detection and counting performance (Table~\ref{tab:detector_robustness}). On the test set, YOLOv5m and RT-DETRv2-S achieve mAP50 of 0.755 and 0.748 and nMAE of 0.080 and 0.091, respectively, while producing nearly identical nMANE ($-0.051$ and $-0.050$). These results indicate that the observed counting behavior is not specific to the CNN-based YOLOv5m detector.

Because RT-DETRv2-S produces differently calibrated confidence scores from YOLOv5m, ByteTrack hyperparameters are selected independently for each detector using validation nMAE.

\begin{table}[t]
\centering
\caption{\textbf{Detector architecture robustness.} Detection and counting performance for YOLOv5m and RT-DETRv2-S on the validation and test sets.}
\label{tab:detector_robustness}
\input{Tables/yolov5_vs_rtdetrv2_performance}
\end{table}

\paragraph{Training Configuration.}
All experiments are trained with YOLOv5m using SGD and the default YOLOv5 training configuration, with a batch size of 16, image size 896, initial learning rate 0.01, momentum 0.937, and weight decay 0.0005. Models are trained on NVIDIA L40S Tensor Core GPUs (8 vCPUs, 64 GiB RAM) on AWS EC2 instances. All other hyperparameters and augmentation settings use YOLOv5 defaults. Raw sonar images are preprocessed to include the background-subtracted frame as an additional channel. This is also the image format used in the dataset release and improves OOD detection performance for the resulting trained model \cite{kay2022caltechfishcountingdataset}.

\paragraph{Tracking and Inference Configuration.}
During inference, detections are produced using the trained YOLOv5m detector with a confidence threshold of 0.1 and a non-maximum suppression (NMS) IoU threshold of 0.25. Tracks are then constructed using ByteTrack, with an IoU association threshold of 0.001, a maximum track age of 20 frames, and a minimum of 5 hits. We performed a ByteTrack hyperparameter sweep over max age (10, 20), min hits (1, 2, 3, 4, 5), and IoU threshold (0.01, 0.001). We found that counting performance was relatively insensitive to these variations. Five locations showed no variation among our hyperparameter sweep (Dungeness, Eel, Kenai-Channel, Kenai-Rightbank, and Nanaimo). Across the full dataset, the average nMAE varied by a relative percentage change of 0.93\% for min hits, 1.39\% for IoU threshold and 4.99\% for max age. The results reported here are from the best performing configuration, selected from the full hyperparameter search. While this does not evaluate alternative trackers, it suggests that the reported directional biases are robust to reasonable variations in ByteTrack hyperparameters.

All inference experiments are executed on NVIDIA L40S Tensor Core GPUs (8 vCPUs, 64 GiB RAM) on AWS EC2 instances, each with 1 TB of attached storage.

\paragraph{Target-River Training.}
For Target River training, Table~\ref{tab:target_location_epochs} shows the epoch scaling used as a function of training-set size. Validation performance is monitored using per-location mAP50, evaluated at intervals corresponding to 10\% of an epoch. Training runs that plateau or saturate early are manually terminated. The best checkpoint is selected using mAP50 aggregated across all per-location validation sets. All experiments use a single run with seed 0.

\begin{table}[ht]
\centering
\caption{Epoch Scaling for Target-Location Training}
\label{tab:target_location_epochs}
\input{Tables/target_location_epochs}
\end{table}

\subsection{Sampling Strategy Selection}
\label{sec:sampling_ablation}
All experiments, both Balanced Sampling (ID) and Balanced Sampling (OOD), use the same underlying balanced sampling procedure with a fixed random seed of 42; the ID/OOD distinction refers to whether the target river is included in training, not to a different sampling design. The per-location training and validation budgets (10k and 3k frames, respectively) were chosen based on the median number of frames containing fish per location in the dataset, ensuring that the sampling target reflects a typical rather than extreme location size. We compared three configurations of this procedure to select the per-location budget: 10k training frames per location with upsampling (locations below the target are upsampled to meet it), 10k without upsampling (available frames only for underrepresented locations), and a lower-budget variant of 1k training frames per location without upsampling, which was terminated early after training stalled with no improvement in validation loss. Validation budgets were 3k, 3k, and 300 frames per location respectively. All three were evaluated by mAP50 and counting metrics.

The 10k no-upsampling configuration achieved higher mAP50 (0.77) than the 10k upsampling configuration (0.74), but the upsampling configuration produced lower counting errors across locations. Because counting accuracy is the operative evaluation criterion and mAP50 is used only for architecture selection, we selected 10k with upsampling for all experiments. This decision itself reflects the paper's metric argument: optimizing for mAP50 would have selected a configuration that performs worse on the metrics practitioners actually care about.

\begin{table}[ht]
\centering
\caption{Sampling strategy comparison. 10k no-upsampling achieves higher mAP50 performance, but 10k-upsampling has lower counting errors.}
\resizebox{\linewidth}{!}{%
    \input{Tables/sampling_ablation}
}
\label{tab:sampling_ablation}
\end{table}

\subsection{Additional Training Conditions}
We evaluate three additional training conditions to test generalization: All Locations (full multi-location coverage), CFC22 Locations, and Kenai Leftbank Only (a single-location setting representative of CFC22's reliance on dominant deployments).

\paragraph{All Locations.} Training splits from each location are combined into a single unified training set; validation splits are combined similarly. This establishes the upper bound when every deployment location contributes training data.

\paragraph{Kenai Leftbank Only.} The model is trained using only Kenai-Leftbank data. We evaluate two settings: (i) validation on Kenai Leftbank, and (ii) cross-bank generalization via validation on Kenai Rightbank. This isolates in-distribution performance and cross-bank generalization within the same river system.

\paragraph{CFC22 Locations.} Training uses only locations present in CFC22: Elwha, all Kenai deployments, and Nushagak, using the updated CFC26 splits.

\begin{table}[ht]
\centering
\caption{Train on Kenai-Leftbank Model performance comparison}
\label{tab:attr4}
\input{Tables/kenai_leftbank_val_diff_4}
\end{table}

\subsection{Additional Generalization Results}

\paragraph{Geographic and ecological coverage both shape generalization.} The CFC22 Locations condition reveals where the original benchmark distribution is sufficient and where it breaks down. On some locations, especially those similar to the CFC22 training locations, this condition performs competitively. The strongest example is Dungeness. Although Balanced Sampling (OOD) and Kenai-Leftbank Only both produce high nMAE of approximately 0.80, CFC22 Locations achieves a much lower nMAE of approximately 0.20. Because Dungeness is geographically close to Elwha, a CFC22 location, this result suggests that the narrower CFC22 training set contains highly relevant signal for Dungeness that is weakened when more ecologically distant locations are added.

This is the main counterexample to a simple “more training diversity is always better” interpretation. For Dungeness, adding additional locations in the Balanced Sampling (OOD) condition appears to dilute the most relevant regional signal rather than improve generalization. In contrast, CFC22 training fails badly at locations that are more distinct from its geographic and ecological scope. At Eel and Mad, CFC22 Locations produces nMAE of approximately 1.0 and 0.73, respectively, indicating near-complete counting failure. Thus, narrow benchmark training can generalize well when the deployment location is similar to the benchmark distribution, but it is unreliable for ecologically distinct rivers.

Chemainus provides a complementary lesson. Target River Only performs worse than Balanced Sampling (OOD), despite being trained on the target location itself, and both conditions are less accurate than Balanced Sampling (ID). This indicates that location-specific data alone is not always sufficient, especially when it is sparse. Broader ecological diversity can provide useful variation that improves counting at the target location, but the best performance comes from combining complementary target-location data with multi-location training rather than relying on either source alone.

Nushagak shows the same principle in the directional setting. Although Nushagak is low-error under most conditions, Kenai-Leftbank Only produces a larger directional error of approximately $-0.65$. This shows that shared Alaskan geography alone is not sufficient for directional generalization; performance improves when training includes broader ecological and geographic diversity.

\subsection{CFC22 vs CFC26}
CFC26 expands Caltech Fish Counting 2022 (CFC22) dataset by including 2.5x as many clips, 2x as many frames, and 2x as many bounding boxes and data from 6 additional deployment locations. Beyond this increase in scale and diversity, CFC26 introduces chronologically ordered, non-overlapping data splits constructed using natural river-specific temporal gaps, held-out-river evaluation for measuring out-of-distribution generalization, directional counting metrics (nMNE and nMANE), and a standardized benchmark designed to better reflect real-world deployment performance.

We additionally use the three-channel representation (raw frame, background-subtracted frame, frame-to-frame difference) introduced in the Caltech Fish Counting Dataset [16] to provide richer temporal context. More targeted approaches, such as synthetic data generation or augmentation strategies designed to better represent sparse, far-range, or low-SNR fish, may help improve robustness to deployment-specific conditions. Exploring such approaches is an important direction for future work but is beyond the scope of the present benchmark and dataset paper.

\subsection{Metric Interpretation and Practitioner Workflow}

In all cases, values closer to zero indicate better performance. An nMAE of 0 indicates that every fish crossing was counted correctly, while a value of 1 indicates that the total number of counting errors equals the number of ground truth crossings, though these could be missed or hallucinated. Likewise, an nMNE of 0 indicates the perfect net-passage estimation, -1 indicates that all true crossings were missed, and +1 shows there were the same number of hallucinated counts as ground truth counts ($2\times$ the count). 

On rivers where fish predominantly travel in a single direction, nMANE and nMNE are often very similar. However, on rivers where different clips will have different dominant directions (e.g., the fish change direction), nMANE measures the undercounts relative to the dominant flow within each clip, where nMNE measures the undercount relative to the dominant direction across all clips. 

Based on informal discussions with fisheries practitioners, it is typical to expect counting errors on the order of 10\% between annotators; this corresponds to an nMANE between -0.1 and 0.1. Most monitoring locations prioritize accurate estimates of net fish population and passage error (captured by nMNE), while others, generally those with lower fish populations or bidirectional fish movement place greater importance on accurately counting individual passages (reflected better by nMAE). When determining whether a model systematically under- or over-counts fish to support calibration, nMANE provides additional insight, especially for rivers with variable crossing directions.

\input{Tables/per_river_counting_errors}

\begin{figure}[t]
  \centering
  \includegraphics[width=\textwidth]{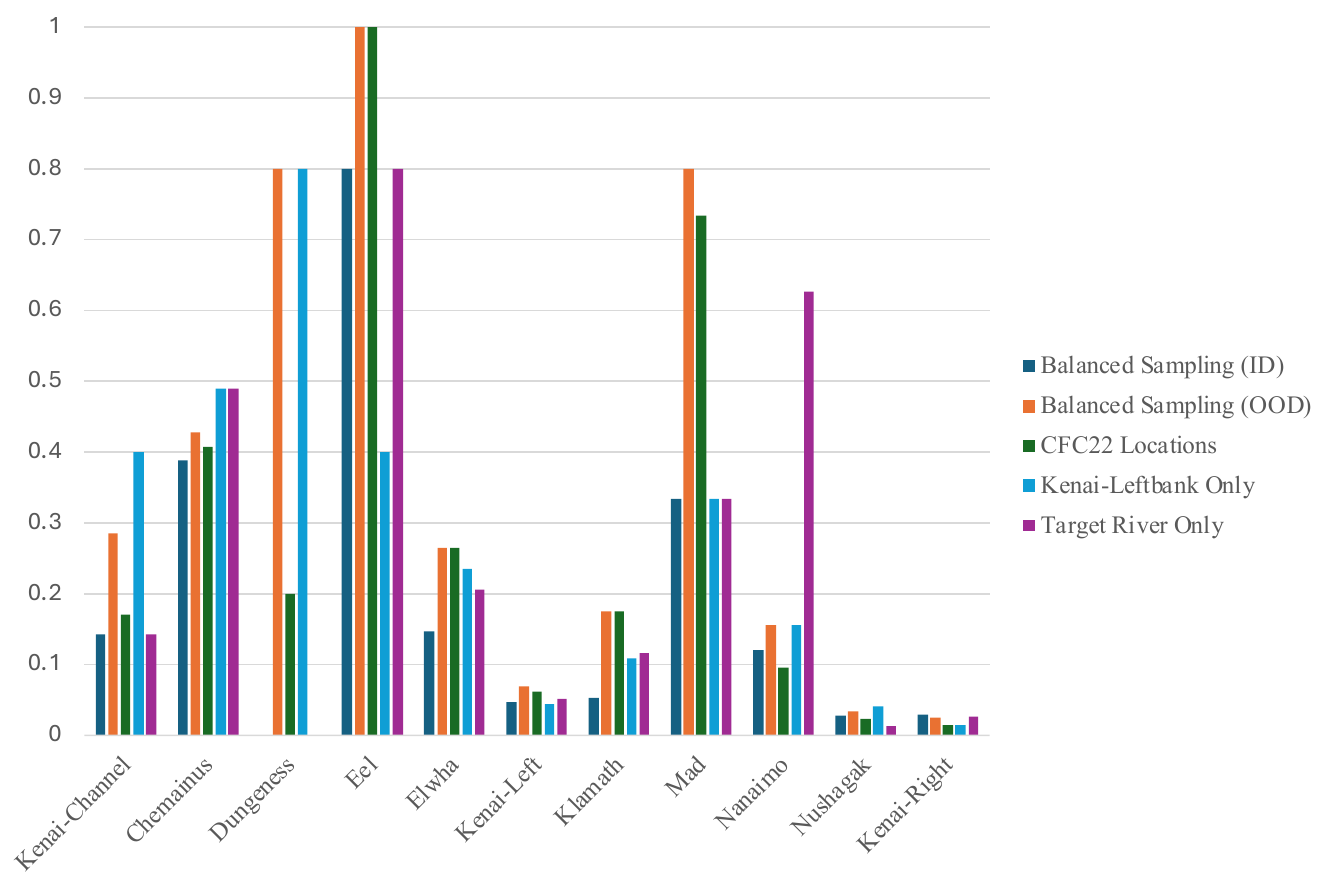}
  \caption{\textbf{Counting error (nMAE) per location across training conditions.} Balanced Sampling (ID), which includes target-location data, achieves the lowest error for most locations, while holding out the target river (OOD) degrades performance substantially at harder locations such as Eel and Dungeness. Training on CFC22 locations alone fails catastrophically on Eel with nMAE $\approx$ 1.0, demonstrating that in-distribution benchmark scores do not predict deployment performance.}
  \label{fig:nmae}
\end{figure}

\begin{figure}[t]
  \centering
  \includegraphics[width=0.7\textwidth]{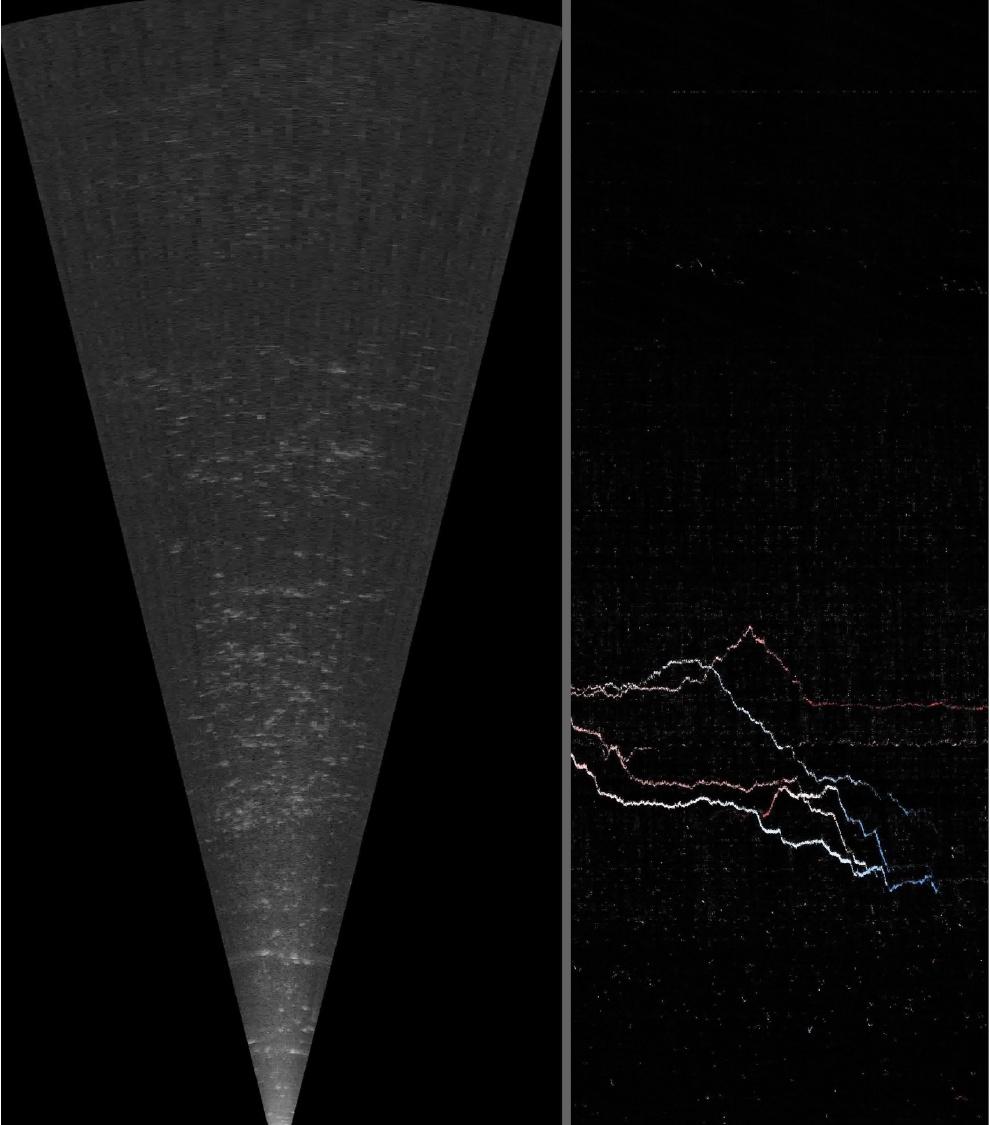}
  \caption{\textbf{Example echogram.} A raw sonar clip (left) and its corresponding echogram (right). While the five fish are barely visible in the noisy, raw imagery, the echogram clearly reveals the tracks over time.}
  \label{fig:echogram}
\end{figure}

\begin{sidewaystable}
\centering
\small
  \caption{Dataset statistics for CFC26 across all deployment locations.
  Each \textit{source video} is an original ARIS file; \textit{clips} are
  contiguous sub-sequences extracted from source videos that contain at least one annotated fish track. Kenai locations are further subdivided into three strata corresponding to distinct sonar mounting positions along the river cross-section. Range window start and stop distances (in meters) are the near and far range boundaries of the sonar sensor as recorded in the ARIS file metadata.}   

\label{tab:dataset_summary}

\resizebox{\textheight}{!}{%

\input{Tables/dataset_full_summary_flipped}
}

\end{sidewaystable}


\newpage
\clearpage

\end{document}

%% file: Tables/dataset_summary_stats_1.tex
\begin{tabular}{lrrrrrrrr}
\toprule
Location & \# Clips & \# Frames & \# Tracks & Track Coverage (\%) & \# Boxes & Avg Track Length \\
\midrule




\textcolor{newlocationblue}{Chemainus}       & 22       & 16,287     & 449        & 41.82               & 8,488      & 22.63            \\
\textcolor{newlocationblue}{Dungeness}       & 6        & 3,118      & 8          & 95.48               & 3,273      & 493.00           \\
\textcolor{newlocationblue}{Eel}             & 27       & 2,900      & 104        & 48.31               & 3,136      & 35.09            \\
Elwha           & 264      & 107,238    & 450        & 31.12               & 4,4703     & 92.38            \\
Kenai-Channel   & 217      & 41,929     & 1,421      & 78.31               & 179,789    & 106.65           \\
Kenai-Leftbank  & 546      & 192,652    & 1,970      & 41.31               & 150,599    & 101.94           \\
Kenai-Rightbank & 656      & 199,109    & 2,430      & 41.80               & 171,003    & 96.95            \\
\textcolor{newlocationblue}{Klamath}         & 2,350     & 518,202    & 3,109     & 45.46               & 249,743    & 78.41            \\
\textcolor{newlocationblue}{Mad}            & 182      & 27,220     & 373        & 79.80               & 27,239     & 82.90            \\
\textcolor{newlocationblue}{Nanaimo}         & 21       & 10,404     & 689        & 77.71               & 59,075     & 110.25           \\
Nushagak        & 72       & 21,528     & 3,069      & 89.80               & 110,869    & 36.41            \\

\midrule
CFC26 Total     & \textbf{4,363}     & \textbf{1,140,587}   & \textbf{14,072}      & \textbf{46.02}               & \textbf{1,007,877}   & 109.86           \\
CFC22 Total     & 1,567     & 527,215    & 8,252       & 42.50               & 515,933    & 96.48 

          \\
\bottomrule
\end{tabular}

%% file: Tables/architecture_selection_3.tex

\begin{tabular}{lcccc}
\toprule
& \multicolumn{2}{c}{YOLOv5} & \multicolumn{2}{c}{YOLOv26} \\
\cmidrule(lr){2-3} \cmidrule(lr){4-5}
Metric & Small & Medium & Small & Medium \\
\midrule
Val mAP@0.5
    & 0.6780 & \textbf{0.6926} & 0.6310 & 0.6610 \\
Val nMAE
    & 0.0729 & \textbf{0.0720} & 0.1857 & 0.1601 \\
Val nMNE
    & \textbf{-0.0325} & -0.0355 & -0.2441 & -0.2003 \\
Val nMANE
    & -0.0372 & \textbf{-0.0338} & -0.1595 & -0.1338 \\
\midrule
Test mAP@0.5
    & 0.7346 & \textbf{0.7546} & 0.6555 & 0.6854 \\
Test nMAE
    & 0.0965 & \textbf{0.0802} & 0.1329 & 0.1029 \\
Test nMNE
    & -0.0649 & \textbf{-0.0397} & -0.1514 & -0.1142 \\
Test nMANE
    & -0.0614 & \textbf{-0.0506} & -0.1104 & -0.0823 \\
\bottomrule
\end{tabular}

%% file: Tables/yolov5_vs_rtdetrv2_performance.tex
\resizebox{\linewidth}{!}{%
\begin{tabular}{lcccccccc}
\toprule
Architecture & Val mAP50 & Val nMAE & Val nMNE & Val nMANE &
Test mAP50 & Test nMAE & Test nMNE & Test nMANE \\
\midrule
YOLOv5m
& \textbf{0.6926}
& \textbf{0.0720}
& -0.0355
& -0.0338
& \textbf{0.7546}
& \textbf{0.0802}
& \textbf{-0.0397}
& -0.0506 \\
RT-DETRv2-S
& 0.6892
& 0.0725
& \textbf{-0.0317}
& \textbf{-0.0316}
& 0.7480
& 0.0906
& -0.0433
& \textbf{-0.0501} \\
\bottomrule
\end{tabular}%
}

%% file: Tables/target_location_epochs.tex
\begin{tabular}{lc}
\toprule
Location      & Epoch \\
\midrule
Chemainus     & 173   \\
Dungeness     & 516   \\
Eel           & 616   \\
Elwha         & 42    \\
Kenai-Channel & 43    \\
Kenai-Left    & 20    \\
Kenai-Right   & 18    \\
Klamath       & 4     \\
Mad           & 63    \\
Nanaimo       & 197   \\
Nushagak      & 81   \\
\bottomrule                                 
\end{tabular}

%% file: Tables/sampling_ablation.tex
\begin{tabular}{lccccccc}
\toprule                                                                                                                                                         
Configuration & Frames & Epochs & Upsample & mAP50 & nMAE & nMNE & nMANE \\
\midrule                                                                                                                                                            
10k, upsampling (selected) & 10k & 74 & \checkmark & 0.74 & 0.07 & -0.04 &  -0.05\\
10k, no upsampling         & 10k & 74 & $\times$   & 0.77 & 0.11 & -0.05 &  -0.07\\
1k, no upsampling          & 1k  & 734 & $\times$   & 0.69 & 0.09 & 0.08  &  -0.05 \\                                                                                              
\bottomrule                                 
\end{tabular}                                                               

%% file: Tables/kenai_leftbank_val_diff_4.tex
\begin{tabular}{lc}
\toprule
Val Location & mAP50 \\
\midrule
Leftbank & 0.80929 \\
Rightbank & \textbf{0.90179} \\
\bottomrule
\end{tabular}

%% file: Tables/per_river_counting_errors.tex
\begin{table*}[t]
\centering
\caption{Per-river counting performance under balanced sampling for in-distribution (ID), balanced sampling out-of-distribution (OOD) evaluation, and on the target river only.}
\label{tab:per_river_counting}
\resizebox{\textwidth}{!}{%
\begin{tabular}{lccc|ccc|ccc}
\toprule
& \multicolumn{3}{c|}{nMAE}
& \multicolumn{3}{c|}{nMANE}
& \multicolumn{3}{c}{nMNE} \\
\cmidrule(lr){2-4}
\cmidrule(lr){5-7}
\cmidrule(lr){8-10}
River
& ID & OOD & Target
& ID & OOD & Target
& ID & OOD & Target \\
\midrule
Nushagak
& 0.028 & 0.035 & 0.014
& -0.028 & 0.028 & -0.014
& -0.028 & 0.028 & -0.014 \\

Kenai-Right
& 0.029 & 0.025 & 0.027
& -0.021 & -0.017 & -0.015
& -0.021 & -0.017 & -0.015 \\

Kenai-Left
& 0.048 & 0.069 & 0.052
& -0.033 & -0.036 & -0.048
& -0.034 & -0.041 & -0.048 \\

Kenai-Channel
& 0.143 & 0.286 & 0.143
& 0.030 & 0.000 & 0.030
& 0.032 & -0.129 & 0.032 \\

Elwha
& 0.147 & 0.265 & 0.206
& -0.094 & -0.219 & -0.094
& -0.115 & -0.269 & 0.038 \\

Klamath
& 0.054 & 0.175 & 0.117
& -0.046 & -0.159 & -0.118
& -0.044 & -0.175 & -0.104 \\

Mad
& 0.333 & 0.800 & 0.333
& -0.027 & -0.108 & -0.081
& -0.448 & -0.828 & -0.379 \\

Nanaimo
& 0.120 & 0.157 & 0.627
& -0.113 & -0.113 & -0.648
& -0.114 & -0.114 & -0.657 \\

Dungeness
& 0.000 & 0.800 & 0.000
& 0.000 & -0.800 & 0.000
& 0.000 & -0.800 & 0.000 \\

Chemainus
& 0.388 & 0.429 & 0.490
& -1.000 & -0.467 & -0.800
& 1.000 & 3.667 & 1.333 \\

Eel
& 0.800 & 1.000 & 0.800
& -0.800 & -1.000 & -0.800
& -0.667 & -1.000 & -1.333 \\
\bottomrule
\end{tabular}%
}
\end{table*}

%% file: Tables/dataset_full_summary_flipped.tex
\begin{tabular}{lccccccccccccccccl}
\hline
 &
  \textbf{Chemainus} &
  \textbf{Dungeness} &
  \textbf{Eel} &
  \textbf{Elwha} &
  \multicolumn{3}{c}{\textbf{Kenai-Channel}} &
  \multicolumn{3}{c}{\textbf{Kenai-Leftbank}} &
  \multicolumn{3}{c}{\textbf{Kenai-Rightbank}} &
  \textbf{Klamath} &
  \textbf{Mad} &
  \textbf{Nanaimo} &
  \textbf{Nushagak} \\ \hline
\textbf{Metric} &
   &
   &
   &
   &
  \textbf{stratum1} &
  \textbf{stratum2} &
  \textbf{stratum3} &
  \textbf{stratum1} &
  \textbf{stratum2} &
  \textbf{stratum3} &
  \textbf{stratum1} &
  \textbf{stratum2} &
  \textbf{stratum3} &
   &
   &
   &
   \\ \hline
\textbf{\# Source Videos} &
  12 &
  6 &
  4 &
  202 &
  44 &
  38 &
  42 &
  251 &
  108 &
  32 &
  113 &
  215 &
  62 &
  980 &
  43 &
  9 &
  11 \\
\textbf{\# Clips} &
  22 &
  6 &
  27 &
  264 &
  84 &
  64 &
  69 &
  362 &
  149 &
  35 &
  176 &
  409 &
  71 &
  2350 &
  182 &
  21 &
  72 \\
\textbf{\# Frames in Clips} &
  16287 &
  3118 &
  2900 &
  107418 &
  16104 & 
  12544 &
  13281 &
  136373 &
  45722 &
  10557 &
  62116 &
  114887 &
  22106 &
  518202 &
  27220 &
  10404 &
  21536 \\
\textbf{\# Frames incl. Track} &
  6811 &
  2977 &
  1401 &
  33416 &
  8144 &
  11459 &
  13231 &
  52523 &
  22156 &
  4899 &
  15646 &
  55262 &
  12312 &
  235550 &
  21721 &
  8085 &
  19334 \\
\textbf{\# Tracks} &
  449 &
  8 &
  104 &
  450 &
  240 &
  457 &
  724 &
  1549 &
  351 &
  70 &
  646 &
  1621 &
  163 &
  3109 &
  373 &
  689 &
  3069 \\
\textbf{Track Coverage (\%)} &
  41.82 &
  95.48 &
  48.31 &
  31.12 &
  50.57 & 
  91.35 &
  99.62 &
  38.51 &
  48.46 &
  46.41 &
  25.19 &
  48.10 &
  55.70 &
  45.46 &
  79.80 &
  77.71 &
  89.78 \\
\textbf{\# Boxes} &
  8488 &
  3273 &
  3136 &
  44703 &
  12387 &
  65214 &
  102188 &
  108656 &
  33703 &
  8200 &
  28681 &
  118894 &
  23428 &
  249743 &
  27239 &
  59075 &
  110869 \\
\textbf{Avg \# Tracks} &
  20.41 &
  1.33 &
  3.85 &
  1.70 &
  2.86 &
  7.14 &
  10.49 &
  4.28 &
  2.36 &
  2.00 &
  3.67 &
  3.96 &
  2.30 &
  1.32 &
  2.05 &
  32.81 &
  42.63 \\
\textbf{Avg Track Length} &
  22.63 &
  493.00 &
  35.09 &
  92.38 &
  47.57 &
  125.18 &
  147.19 &
  81.09 &
  103.83 &
  120.89 &
  56.32 &
  90.47 &
  144.07 &
  78.41 &
  82.90 &
  110.25 &
  36.41 \\
\textbf{Median BBox Width (px)} &
  44.56 &
  41.19 &
  44.86 &
  68 &
  68 &
  64.64 &
  80.02 &
  70 &
  66 &
  84 &
  63 &
  79 &
  82 &
  56.96 &
  32.55 &
  34.37 &
  99.00 \\
\textbf{Median BBox Height (px)} &
  18.00 &
  9.61 &
  15.56 &
  23 &
  18.83 &
  17.74 &
  22 &
  20 &
  21 &
  24 &
  17 &
  21 &
  25 &
  16.54 &
  11.49 &
  10.91 &
  31.00 \\
\textbf{Median BBox Area (px$^2$)} &
  805.01 &
  395.35 &
  675.51 &
  1620.00 &
  1295.05 &
  1144.01 &
  1743.69 &
  1368.00 &
  1386.00 &
  1990.00 &
  1089.99 &
  1653.00 &
  2072.00 &
  921.07 &
  366.94 &
  378.78 &
  3120 \\
\textbf{Relative BBox Size} &
  1.03E-03 &
  5.68E-04 &
  2.03E-04 &
  2.05E-03 &
  1.23E-02 &
  3.24E-03 &
  2.25E-03 &
  3.15E-03 &
  9.19E-04 &
  1.15E-03 &
  6.09E-03 &
  2.06E-03 &
  1.21E-03 &
  2.62E-04 &
  2.46E-04 &
  5.83E-04 &
  1.35E-03 \\
\textbf{Avg Window Start (meters)} &
  0.68 &
  1.42 &
  0.77 &
  0.67 &
  2.50 &
  6.00 &
  11.99 &
  3.50 &
  8.00 &
  21.93 &
  3.50 &
  8.00 &
  22.00 &
  0.68 &
  0.69 &
  1.68 &
  0.70 \\
\textbf{Avg Window Stop (meters)} &
  12.46 &
  21.66 &
  42.24 &
  15.16 &
  5.99 &
  11.99 &
  19.99 &
  8.79 &
  21.98 &
  32.99 &
  7.99 &
  19.02 &
  32.99 &
  33.62 &
  50.75 &
  29.69 &
  10.09 \\
\textbf{Avg Pixel Meter Size} &
  9.53E-03 &
  1.73E-02 &
  1.58E-02 &
  1.17E-02 &
  7.25E-03 &
  7.32E-03 &
  7.39E-03 &
  7.28E-03 &
  7.31E-03 &
  1.13E+01 &
  7.27E-03 &
  7.32E-03 &
  7.48E-03 &
  1.23E-02 &
  2.78E-02 &
  2.43E-02 &
  4.43E-03 \\
  \textbf{Upstream Direction} &
  Left-to-Right & 
  Left-to-Right & 
  Left-to-Right & 
  Left-to-Right & 
  Left-to-Right & 
  Left-to-Right & 
  Left-to-Right & 
  Left-to-Right & 
  Left-to-Right & 
  Left-to-Right & 
  Right-to-Left & 
  Right-to-Left & 
  Right-to-Left & 
  Right-to-Left & 
  Right-to-Left & 
  Right-to-Left & 
  Right-to-Left 
\end{tabular}